\documentclass[letterpaper]{article} 
\usepackage{aaai24}  
\usepackage{times}  
\usepackage{helvet}  
\usepackage{courier}  
\usepackage[hyphens]{url}  
\usepackage{graphicx} 
\usepackage{natbib}  
\usepackage{caption} 
\usepackage{algorithm}
\usepackage{algorithmic}

\usepackage{amsmath}
\usepackage{amssymb}
\usepackage{stmaryrd}

\usepackage{newfloat}
\usepackage{listings}
\DeclareCaptionStyle{ruled}{labelfont=normalfont,labelsep=colon,strut=off} 
\floatstyle{ruled}
\newfloat{listing}{tb}{lst}{}
\floatname{listing}{Listing}
\title{Automated Process Planning Based on a Semantic Capability Model and SMT}

\author {
    Aljosha Köcher\equalcontrib,
    Luis Miguel Vieira da Silva\equalcontrib,
    Alexander Fay
}
\affiliations {
    Institute of Automation, Helmut Schmidt University, Hamburg, Germany\\
    \{aljosha.koecher, miguel.vieira, alexander.fay\}@hsu-hh.de
}

\begin{document}

\maketitle

\begin{abstract}
In research of manufacturing systems and autonomous robots, the term \emph{capability} is used for a machine-interpretable specification of a system function. 
Approaches in this research area develop information models that capture all information relevant to interpret the requirements, effects and behavior of functions. These approaches are intended to overcome the heterogeneity resulting from the various types of processes and from the large number of different vendors.
However, these models and associated methods do not offer solutions for automated process planning, i.e. finding a sequence of individual capabilities required to manufacture a certain product or to accomplish a mission using autonomous robots. 
Instead, this is a typical task for AI planning approaches, which unfortunately require a high effort to create the respective planning problem descriptions.
In this paper, we present an approach that combines these two topics: Starting from a semantic capability model, an AI planning problem is automatically generated. 
The planning problem is encoded using \emph{Satisfiability Modulo Theories} and uses an existing solver to find valid capability sequences including required parameter values. 
The approach also offers possibilities to integrate existing human expertise and to provide explanations for human operators in order to help understand planning decisions.
\end{abstract}

\section{Introduction}  
\label{sec:introduction}
Today's companies operate in a market environment that is defined by rapidly changing customer demands, resulting in shorter product life cycles and smaller batch sizes. 
To succeed in this environment, companies must be able to adjust their manufacturing systems quickly to changing requirements.
Users of autonomous robots trying to accomplish complex missions with different robots and their functions face similar challenges. In the context of both manufacturing as well as autonomous systems, the right systems and functions must be selected and modified if necessary.
The ability of systems to do this is often referred to as \emph{changeability} \cite{WEN+_ChangeableManufacturingClassificationDesign_2007b}.
Changeability covers both changes on a physical as well as on a logical / control level \cite{ElM_ReconfigurableProcessPlansFor_2007}.

When looking at changeability on control level, process planning and replanning must be performed more frequently as a consequence of decreasing lot sizes, changing system configurations and different use cases. 
Hence, to remain efficient and avoid delays, process planning can no longer be done manually. 
Instead, methods for automated process planning incorporating all data and knowledge are needed \cite{WBS+_AnOntologybasedMetamodelfor_9820209112020}.
These methods include two aspects: On the one hand, machine-interpretable models are needed that can represent all data and knowledge needed for process planning and control. 
And on the other hand, an encoding of a planning problem based on these models needs to be established in a suitable formalism.
While there is a variety of approaches for either one of these aspects, no approach properly integrates both aspects \cite{KHW+_AResearchAgendafor_2022}. 

For the first aspect, there are approaches focusing on a description of relevant knowledge and machine functions as so-called \emph{capability} models, but such models and their associated methods, e.g., automated reasoning, typically do not contain methods for robust process planning.
Because ontologies --- typically modeled using the Web Ontology Language (OWL) --- offer a number of advantages for capturing relevant knowledge, they are often considered for defining these models \cite{WBS+_AnOntologybasedMetamodelfor_9820209112020}.

For the second aspect, approaches focusing on planning require (manual) effort to create the necessary planning encoding and do not satisfy the requirements of changeable systems \cite{RoNi_Automatedprocessplanningfor_2017}.
Modern approaches to AI planning make use of \emph{Satisfiability Modulo Theories} (SMT) and corresponding solvers \cite{CMZ_PlanningforHybridSystems_2020}.

\begin{figure}[htb]
    \centering
    \includegraphics[width=\linewidth]{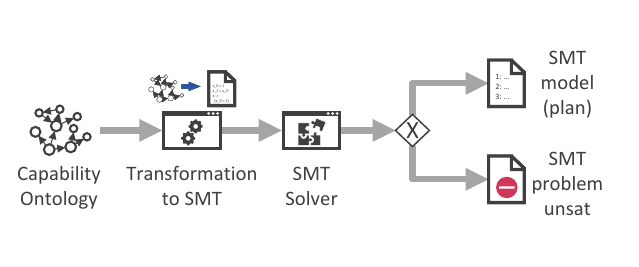}
    \caption{Overview of the proposed planning approach. Required and provided capabilities in an ontology are transformed into an SMT planning problem. If the problem is satisfiable, the resulting model is a valid process plan.}
    \label{fig:concept}
\end{figure}

With this contribution, we aim for a combination of these two aspects: We present an approach to automatically generate a planning problem in Satisfiability Modulo Theories (SMT) starting from an existing semantic capability model. For this capability model, an engineering method supporting model development already exists and no additional effort is required for process planning.
The generated SMT problem can be solved using off-the-shelf SMT solvers. If the problem is satisfiable, the resulting SMT model is a valid plan consisting of capabilities.
The overall approach is shown in Figure~\ref{fig:concept}.

The remainder of this paper starts with an overview of related work in the fields of capability modeling and process planning in Section~\ref{sec:relatedWork}.
Section~\ref{sec:capabilityModeling} explains the fundamental elements of our existing capability model. Introducing these fundamentals is necessary in order to understand the actual planning approach, which is presented in Section~\ref{sec:planningApproach}. Since this paper presents early results and work that is still in progress, there is no evaluation yet. Instead, this paper ends with a conclusion and an outlook on our next steps.

\section{Related Work}
\label{sec:relatedWork}
In this section, existing approaches to both previously mentioned aspects of an integrated process planning approach are discussed. 
First, related work in the area of capability modeling is analyzed before looking into existing process planning methods.

One of the earliest approaches that use the term \emph{Capability} in the context of an information model for manufacturing functions is \cite{AmDu_AnUpperOntologyfor_2006}. In this paper, \citeauthor{AmDu_AnUpperOntologyfor_2006} present \emph{Manufacturing Service Description Language} (MSDL), an upper ontology that contains core concepts and relations to describe manufacturing functions. MSDL allows to describe constraints
of capabilities such as physical attributes of inputs and outputs.
This early approach was followed by a variety of others, in which ontologies are often used to describe the complicated relationships underlying manufacturing processes in a semantic way. 

In \cite{JSL_FormalResourceandCapability_2016}, the authors introduce a formal capability ontology that can be used to support automated reconfiguration of machines. In \cite{JSL_FormalResourceandCapability_2016}, capabilities are modeled as individuals of an extensive ontology that also covers resources and products. Complex capabilities can be composed of multiple simple ones.

Another approach also using an ontology is presented in \cite{GGR_SupportingSkillbasedFlexibleManufacturing_2020}. In this approach, capabilities are modeled as classes and OWL class expressions are used to formulate restriction (e.g., about admissible input parameters).

A disadvantage of all these approaches is that the developed ontologies are not made available and that they are hardly reusable, because they were often developed for certain use cases.
In \cite{KHV+_AFormalCapabilityand_2020}, we present a capability ontology that is based on standards, modular and extensible and that is continually maintained and extended. Details of this ontology are introduced in Section~\ref{sec:capabilityModeling}, as the planning approach presented in this paper is based on this ontology.

\medskip

The second aspect, automated process planning, has been studied for decades as a field of symbolic AI. Classical AI planning is concerned with "finding a sequence of actions to accomplish a goal in a discrete, deterministic, static and fully observable environment" \cite{RuNo_Artificialintelligence_2021}.
To express planning tasks in a standardized way, Planning Domain Definition Language (PDDL) was introduced by \cite{GHK+_PDDLThePlanningDomain_1998}.
In PDDL, domains can be expressed in a reusable way with all available actions and the preconditions as well as effects that govern applicability of these actions. Specific problems with a given initial and desired goal state are modeled separately and refer to a domain definition \cite{GHK+_PDDLThePlanningDomain_1998}.
While original PDDL allowed only boolean propositions to be used, multiple extensions to the language were created to cover, e.g., planning with numeric values and temporal planning \cite{FoLo_PDDL2.1:AnExtension_2003}, or planning with probabilistic effect \cite{YoLi_PPDDL1.0:TheLanguage_2004}.
PDDL is used for various planning tasks in manufacturing, e.g., to find plans for a fleet of autonomous robots \cite{CrPe_Missionplanningfora_2015} or in kit-building for assembly \cite{KSL+_Towardsrobustassemblywith_2015}.

There are a few approaches that combine the two aspects of system / capability modeling with automated process planning. 
The contribution in \cite{Michniewicz.2016} considers how a plan can be automatically generated from a product to be assembled. 
This is done by automatically creating a virtual representation of a product from a CAD file. 
There is also a virtual representation of the devices in a robotic cell. Finally, the required and offered capabilities, called functional primitives, are compared. 
However, the virtual representation of the functional primitives is created for this specific use case and there is no reusable semantic model.
Furthermore, the planning approach presented in \cite{Michniewicz.2016} is tailored to assembly processes and does not use a domain independent language like PDDL. Instead, a production graph is created in which devices and their functions are connected if they can be combined or executed one after the other. 

In \cite{GGR_SupportingSkillbasedFlexibleManufacturing_2020}, capability descriptions are transferred from their ontological representation to PDDL. For this purpose, however, the model must have certain class assignments from a PDDL ontology, which tightly couples the model and the PDDL planning description.
\citeauthor{WVN+_FlexibleProductionSystems:Automated_2019} generate PDDL domain and goal files from system models defined according to ISA 95 \cite{WVN+_FlexibleProductionSystems:Automated_2019}.

While approaches like \cite{GGR_SupportingSkillbasedFlexibleManufacturing_2020} and \cite{WVN+_FlexibleProductionSystems:Automated_2019} are valuable in reducing the effort required to manually create the textual PDDL files, they suffer from one major drawback: PDDL is not expressive enough to cover real-world applications \cite{RoNi_Automatedprocessplanningfor_2017}.

Instead, formalizing planning as a problem of satisfiability has emerged as a competitor to PDDL in recent years, in part because SAT solvers have made significant progress.
A first work to formalize planning as a satisfiability problem is presented in \cite{KaSe_PlanningasSatisfiability_1992}, in which \citeauthor{KaSe_PlanningasSatisfiability_1992} define a planning encoding as a boolean SAT problem.
More expressive than boolean SAT is the use of SMT, a formalism that extends SAT with so-called background theories. With the help of these theories, other data structures such as reals or arrays can be considered in addition to boolean propositions \cite{Cla_SatisfiabilityModuloTheories_2021}.
Much research has been done in recent years on the use of SMT for encoding planning problems \cite{BLP+_SMTbasedPlanningforRobots_2019, BGM+_SMTBasedNonlinearPDDL+Planning_2015, CMZ_PlanningforHybridSystems_2020}. 
One approach that has received particular attention is \cite{CMZ_PlanningforHybridSystems_2020}, in which an automatic mapping of the entire PDDL+ language into SMT is provided in order to solve PDDL+ planning problems using powerful SMT solvers.

Formulating a planning problem directly as a satisfiability problem in SMT has a number of advantages compared to using a dedicated planning language such as PDDL.
For example, plans with parallel actions can be efficiently implemented with a suitable SMT encoding \cite{CMZ_PlanningforHybridSystems_2020}.
Furthermore, in addition to actions, any additional information, such as human expert knowledge in the form of known or forbidden (partial) solutions, can be integrated. If no plan can be found for a problem, a counterexample can be generated and the root cause for non-satisfiability can be found, which helps to make the planning approach comprehensible for human users.
And when it comes to solving a problem, the performance of modern SMT solvers far exceeds that of most non-SMT-based PDDL solvers \cite{CMZ_PlanningforHybridSystems_2020}. 

Nevertheless, formulating a planning problem directly as a satisfiability problem in SMT is challenging and much more difficult than using a dedicated language such as PDDL.
Therefore, the goal of this contribution is to present an automated transformation of a capability ontology into a planning problem encoded as a satisfiability problem in SMT, so that the time-consuming manual creation of the planning problem is no longer necessary.

\section{Capability Modeling}  
\label{sec:capabilityModeling}
In \cite{KHV+_AFormalCapabilityand_2020}, we present an ontology\footnote{https://github.com/caskade-automation/cask} to describe machines, their capabilities and executable skills. This ontology is composed of different so-called \emph{Ontology Design Patterns} (ODPs). Every ODP is a self-contained ontology with a defined purpose and based on a standard, e.g., for representing the structure of a resource. 
All ODPs can be individually reused and thus benefit from improvements in various projects and contexts.
In order to obtain a coherent capability ontology, the ODPs are imported into a new ontology and relations between elements of the ODPs are defined.
Of this ontology, only the capability aspect is relevant for this paper. Therefore, an overview of the capability aspect is given, while other aspects are not explained here.

A capability is a specification of a function that is directly related to a (manufacturing) process \cite{KBH+_AReferenceModelfor_15.09.2022b}. We model processes according to \cite{VDI_36821_Formalisedprocessdescriptions}. A process consists of process operators that can have products, information or energy as inputs and outputs. These inputs and outputs are further characterized by properties which are modeled using so-called data elements according to \cite{IEC_61360_Standarddataelementtypes}. A data element consist of a type description and instance description. While a type description contains type-related information about a property (e.g., ID, name, unit of measure), an instance description captures one distinct value of that property.
Instance descriptions can be further subdivided by their expression goal into requirements, assurances and actual (i.e., measured) values.
We assume that type descriptions are unique in our ontology and are ideally taken from a classification system such as ECLASS\footnote{https://eclass.eu/en}.
Data elements are directly linked to their type and instance descriptions and are also bound to the respective carrier of the property (e.g. a product, a resource or other model elements). 

\begin{figure*}
    \centering
    \includegraphics[width=0.9\linewidth]{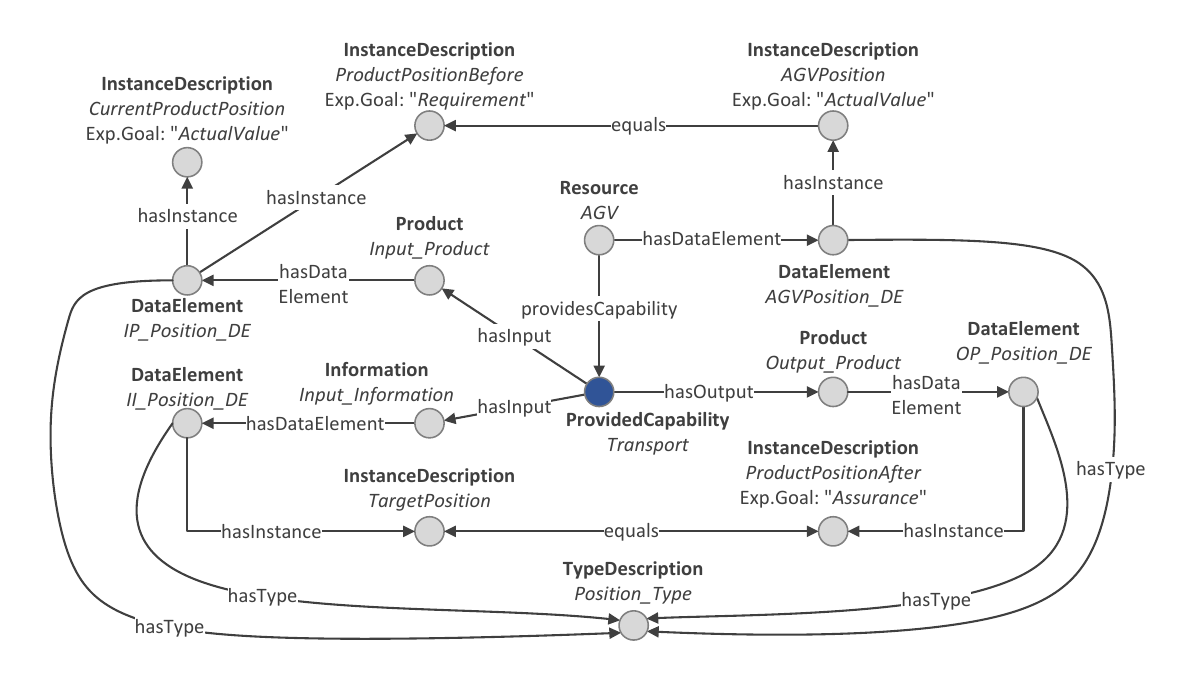}
    \caption{Simplified representation of a transport capability and its properties modeled with the CaSk ontology.}
    \label{fig:capModel}
\end{figure*}

With this approach, capabilities can be modeled with their possible inputs and outputs in a detailed way.
In Figure~\ref{fig:capModel}, a simplified excerpt of an ontology including a transport capability is shown. This capability moves a product (\verb|Input_Product|) from its current position (\verb|CurrentProductPosition|) to a target position (\verb|TargetPosition|).

For properties of inputs, requirements of permissible values can be expressed. These requirements can either be static or dependent on other model elements. For example, a drilling capability could only drill holes up to a fixed maximum diameter of 50 mm. 
In the example shown in Figure~\ref{fig:capModel}, the current position of the input product is required to be equal to the position of the transport resource (\verb|AGVPosition|), as otherwise the product cannot be picked up.

In addition to requirements, unbound parameters can also be modeled. These parameters are selected by a user or determined during planning. In Figure~\ref{fig:capModel}, for example, \verb|TargetPosition| is an unbound parameter for which no requirements exist.

The properties of outputs are typically modeled as assurances. Here, too, static values as well as values depending on other elements can be represented. In Figure~\ref{fig:capModel}, the \emph{Transport} capability guarantees that the assured position after transport is equal to the desired position to be selected (i.e., the input parameter \verb|TargetPosition|).
Figure~\ref{fig:capModel} only shows equality relations for simplicity purposes. In our model, arbitrary mathematical relations can be expressed using the \emph{OpenMath} ontology presented in \cite{Wen_OpenMathRDF:RDFencodingsfor_2021}.

For real capabilities, the resulting ontology can quickly become complex. It is therefore important to understand that this ontology does not have to be modeled in a textual way. Instead, we previously presented engineering methods that can be used to generate a capability ontology efficiently and in a semi-automatic way \cite{KHC+_AutomatingtheDevelopmentof_9820209112020, VKT+_APythonFrameworkfor_2023}.

\section{Capability Planning Approach}
\label{sec:planningApproach}
This section describes a planning approach for capabilities encoded using SMT, which strongly builds on and extends the planning approach presented in \cite{CMZ_PlanningforHybridSystems_2020}. 
In \cite{CMZ_PlanningforHybridSystems_2020}, the authors generate an efficient SMT encoding for a PDDL+ problem and solve the encoding with an SMT solver in order to find a plan. 
In our approach, SMT is generated directly from a capability ontology. 
Since the concepts of PDDL and our capability model are similar, the approach in \cite{CMZ_PlanningforHybridSystems_2020} can be used as a basis.

The main elements of PDDL are actions, their preconditions and effects, as well as a description of the initial and goal state. This is comparable to the capabilities and their inputs and outputs within the capability model. 
Theoretically, a PDDL encoding could be generated from the capability model first before the approach in \cite{CMZ_PlanningforHybridSystems_2020} could be used to solve the planning problem. 
However, the capability ontology contains additional information such as constraints between properties and capabilities that are not covered by \cite{CMZ_PlanningforHybridSystems_2020}. 
A direct transformation of this information is more comprehensive, since PDDL as an additional level of abstraction is skipped. 
Furthermore, a direct transformation is more efficient as it omits an unnecessary transformation step. 

Figure~\ref{fig:concept} gives an overview of the procedure. 
Provided capabilities, i.e., capabilities provided by resources, can be compared to PDDL actions and form the planning domain. 
Required capabilities and their properties define the problem to be solved. 
Provided and required capabilities are used to generate an SMT planning encoding similar to \cite{CMZ_PlanningforHybridSystems_2020}.
An SMT solver, in our case the Z3 solver initially presented in \cite{MoBj_Z3:AnEfficientSMT_2008}, is used to solve the generated planning encoding.  
If a solution is found, a corresponding plan with a sequence of capabilities and their properties can be returned. 

We first explain the differences between our approach and that of \citeauthor{CMZ_PlanningforHybridSystems_2020} before introducing the SMT encoding and then presenting the implementation that generates the SMT planning encoding from the capability model. 

One difference between \cite{CMZ_PlanningforHybridSystems_2020} and our approach is that, for now, we are not considering processes and events according PDDL+. 
In this contribution we only consider capabilities that are similar to instantaneous actions in PDDL. 
Processes and events are ultimately actions that are executed immediately when their preconditions are met. 
A process runs until its condition is no longer met, while an event is an instantaneous action. 
These two concepts are not yet included in the capability model. 

Another difference is that --- unlike PDDL and therefore also unlike \cite{CMZ_PlanningforHybridSystems_2020} --- capability models do not have a concept of global variables. 
Capabilities may be described independently of each other by different stakeholders, e.g. system vendors. 
As a consequence, different terms --- or IRIs in the case of an ontology --- may be used to describe the same object. As an example, consider a blank part that can theoretically be used as an input to multiple capabilities. 
In the same way, one physical object referred to with different IRIs may be passed between successive capabilities, so changes to such an object under all its IRIs must be kept consistent. 
OWL ontologies can natively handle such situations as there is no unique name assumption and instead, multiple terms can refer to one concept. 
However, in SMT this situation has to be explicitly taken care of. In this  approach, products of different capabilities with different IRIs but the same product type are referred to as \emph{synonymous products}. Properties of a product $A$ and a synonymous product $B$ with the same property type description are called \emph{synonymous properties}. 
Based on these definitions, two capabilities with properties synonymous to each other are defined as \emph{synonymous capabilities}.

Another difference to the contribution of \cite{CMZ_PlanningforHybridSystems_2020} is that, in addition to conventional preconditions and effects with specific values or simple relations, so-called \emph{capability constraints} can be formulated. Capability constraints are mathematical expressions forming a relation between properties. 
In the example of Figure~\ref{fig:capModel}, the \verb|equals| relation between \verb|TargetPosition| and \verb|ProductPositionAfter| is a simplified depiction of a capability constraint. As described in Section~\ref{sec:capabilityModeling}, arbitrary mathematical expressions can be modeled using \emph{OpenMath}.

\subsection{SMT Encoding}
\label{subsec:encoding}
An SMT encoding for capability planning is defined, which  declares variables for capabilities and properties on the one hand, and formulates constraints based on the capability description that enforce certain changes of the variables on the other hand. 
If all constraints are satisfiable by assigning specific values to the variables, this solution can be considered a valid plan of the capability planning problem.

We make use of the notion of \emph{happenings} as introduced by \cite{CMZ_PlanningforHybridSystems_2020}. The planning problem considers different points in time $t$ from a fixed number of points in time $n$, so-called \emph{happenings}, which are defined as "a time-stamped moment of discrete change" \cite{CMZ_PlanningforHybridSystems_2020}. 
In a happening, discrete changes can result from the application of capabilities. 
Thus, for each happening, it is determined whether a capability can be applied at that moment and what values the properties have at that moment.

At the beginning, variables are declared that represent the capabilities and their properties. 
For this, we consider the set of provided capabilities $C$, and the set of properties $Q$, which contains all data elements associated with a resource, product or information related to a capability.
Properties are distinguished into boolean and real properties to define corresponding variables in SMT. 
Accordingly, there is a subset $P$ for boolean propositions and a subset $R$ for real variables.
\begin{equation}
	P \subseteq Q \quad and \quad R \subseteq Q 
\end{equation}

In this SMT encoding, a happening consists of only two layers compared to the approach in \cite{CMZ_PlanningforHybridSystems_2020} because events are not yet considered. 
In every happening $t$, layer $0$ contains the initial state of all variables before capabilities are applied. 
After applying one or more capabilities in a happening, layer $1$ contains the final state of all variables after the effects of applied capabilities have been realized. 

Therefore, each happening $x_t$ is defined as a tuple of properties $q \in Q$ for the two layers $0$ and $1$ and capabilities $c \in C$: 
\begin{equation}
	x_t := \{Q_{t,0}, Q_{t,1}, C_t\}
\end{equation}

\begin{figure}[htb]
    \centering
    \includegraphics[width=\linewidth]{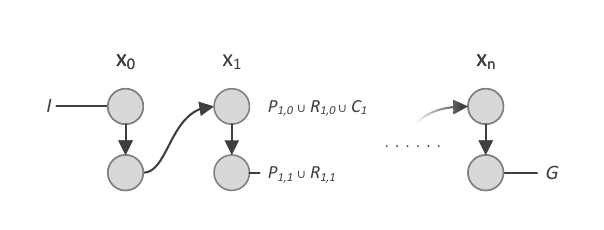}
    \caption{Encoding a capability plan as a sequence of happenings. Adapted from \cite{CMZ_PlanningforHybridSystems_2020}}
    \label{fig:happening}
\end{figure}

Figure~\ref{fig:happening} demonstrates the encoding of a plan as a bounded number of happenings. 
The initial state $I$ is set to happening $0$ and layer $0$ before any application of a capability. 
When capabilities are applied, the effects are set to the variables in layer $1$. 
Corresponding happenings are sequenced until the goal $G$ is eventually reached, in which all property variables have to take the values defined by a required capability and its desired property values.

In the following, the encoding is explained on the basis of one happening. 
For more happenings, variables and constraints are declared analogously with an incremented index corresponding to the number of happenings. 

\subsubsection{Synonymous Properties and Synonymous Capabilities}
\label{subsubsec:synProps&Caps}
Properties of resources, products, or information are changed by the application of capabilities. 
A property can not only be changed by a capability its directly related to, but also by a synonymous capability.  
Therefore, synonymous capabilities must also be included in the following constraints.
For this purpose, first the set $Q^{syn}_q$ is defined, which contains all synonymous properties of a property $q$: 
\begin{equation}
	Q^{syn}_q \subset Q
\end{equation}
Capabilities are synonymous with respect to a certain property if a property of another capability is synonymous to the property considered.
The set $C^{syn}_q$ contains all synonymous capabilities with respect to a certain property $q$, i.e., capabilities that this property is not directly related to, but that have an influence on this property through one of its synonymous properties. 
\begin{equation}
	C^{syn}_q \subset C 
\end{equation}

\subsubsection{Proposition Continuation across Layers}
\label{subsubsec:propositionSupport}
Just like \cite{CMZ_PlanningforHybridSystems_2020}, the encoding needs to contain constraints to ensure that variables remain consistent across layers of each happening. 
First, we consider the propositions that represent boolean variables. 
The two possible values of a proposition, i.e., true or false, can be distinguished. 
A proposition can only change between layers if a capability is applied that has an effect on this proposition.
In analogy to \cite{CMZ_PlanningforHybridSystems_2020}, we formulate the set of $\mathit{eff}_c$ that contains all directly affected properties of a capability $c$.
A capability has a direct effect on a property if this property is part of the output of the capability, where positive and negative effects need to be distinguished.
To distinguish between a positive and a negative effect of a capability on a proposition, two further sets emerge:
\begin{equation}
	\mathit{eff}^+_c \subseteq \mathit{eff}_c \quad and \quad \mathit{eff}^-_c \subseteq \mathit{eff}_c 
\end{equation}

With these definitions, proposition continuation across layers can be defined as follows.

A proposition is true in layer $1$ only if it was already true in layer $0$, or if a capability with a direct or indirect positive effect on that proposition has been applied. 
\begin{equation}
    \label{eqn:propCont_pos}
	\bigwedge_{p \in P} p_{t,1} \implies
        (p_{t, 0} \vee \bigvee_{c \in C^+_p} c_t \vee \bigvee_{c^{syn} \in C^{syn, +}_p} c^{syn}_t)
\end{equation}
In Equation~\ref{eqn:propCont_pos}, $C^+_p$ is the set of capabilities with a direct positive effect on $p$. 
\begin{equation}
    \label{eqn:posCap}
    C^+_p \subset C \mid p \in \mathit{eff}^{+}_{c}
\end{equation}
And $C^{syn,+}_p$ is the set of synonymous capabilities of $c$ with regards to property $p$ that have a positive effect on a synonymous property $p^{syn}$ of p:
\begin{equation}
    \label{eqn:posSynCap}
    C^{syn, +}_p \subseteq C^{syn}_p \mid p^{syn} \in Q^{syn}_p \wedge p^{syn} \in \mathit{eff}^{+}_{C^{syn}_p}
\end{equation}

Similarly, a proposition is false in layer $1$ only if it was already false in layer $0$, or if a capability with a negative effect on that proposition has been applied. 
\begin{multline}
	\label{eqn:propCont_neg}
	\bigwedge_{p \in P} \lnot p_{t,1} \implies
        (\lnot p_{t, 0} \vee \bigvee_{c \in C^-_p} c_t \vee \bigvee_{c^{syn} \in C^{syn, -}_p} c^{syn}_t)
\end{multline}
Analog to Equation~\ref{eqn:propCont_pos}, in Equation~\ref{eqn:propCont_neg}, $C^-_p$ is the set of capabilities with a direct negative effect on $p$ and is defined analogously to $C^+_p$ in Equation~\ref{eqn:posCap}. And $C^{syn,-}_p$ is the set of synonymous capabilities of $c$ with regards to property $p$ that have a negative effect on a synonymous property $p^{syn}$ of $p$. $C^{syn,-}_p$ is defined analogously to $C^{syn,+}_p$ in Equation~\ref{eqn:posSynCap}.

If the effect of a capability only requires a proposition to remain the same, there is no change in the proposition between layers and the capability is not considered for the constraints defined in Equations~\ref{eqn:propCont_pos} and \ref{eqn:propCont_neg}.

\subsubsection{Real Variable Continuation across Layers}
In addition to boolean propositions $P$, real variables $R$ must also be restricted so that unwanted change between layers is prohibited. 
Analogous to proposition continuation across layers, constraints are defined to keep real variables consistent between the two layers. 
In all happenings, a real variable must have the same value in layer $1$ as in layer $0$ if no capability with a numerical effect on this real variable is applied. 
For this purpose, the subset of capability effects $\mathit{eff}_c^{num}$ is defined, which contains all real variables on which the considered capability has a numerical effect. 
\begin{equation}
	\mathit{eff}^{num}_c \subseteq \mathit{eff}_c  
\end{equation}
Also considered for the constraint are all synonymous capabilities that have a numerical influence on a synonymous property. 
This results in the following constraint: 
\begin{equation}
    \label{eqn:realVariableContinuation}
	\bigwedge_{r \in R}(\bigwedge_{c \in C^{num}_r} \neg c_t \wedge 
 \bigwedge_{c^{syn} \in C^{syn, num}_r} \neg c^{syn}_t)
    \implies (r_{t,1} = r_{t, 0})
\end{equation}

In Equation~\ref{eqn:realVariableContinuation},  $C^{num}_r$ is the set of capabilities with a direct numerical effect on $r$. 
\begin{equation}
    \label{eqn:numCap}
    C^{num}_r \subset C \mid r \in \mathit{eff}^{num}_{c}
\end{equation}
And $C^{syn, num}_r$ is the set of synonymous capabilities of $c$ with regards to property $r$ that have a numerical effect on a synonymous property $r^{syn}$ of $r$:
\begin{equation}
    \label{eqn:numSynCap}
    C^{syn, num}_r \subseteq C^{syn}_r \mid r^{syn} \in Q^{syn}_r \wedge r^{syn} \in \mathit{eff}^{num}_{C^{syn}_r}
\end{equation}


\subsubsection{Capability Preconditions and Effects}
Next, capabilities and the constraints governing their applicability must be encoded in SMT. 
A capability can only be applied if all its preconditions are met. 
These preconditions are derived from input properties of capabilities with an expression goal of \emph{requirement}. 
If the preconditions of a capability $c$ are met for any happening in layer $0$, the capability can be applied so that its effect changes the property values in layer $1$. 
Applying a capability causes its boolean capability variable to be set to true.
\begin{equation}
    \label{eqn:capabilityPrecondition}
	\bigwedge_{c \in C} c_t \implies pre_c(Q_{t,0})
\end{equation}
The function $pre_c$ returns all input properties that are a requirement and thus a precondition for the application of a capability. 

If a capability is applied, all effects must be realized, i.e., output properties in layer $1$ must be set according to their assured values. 
\begin{equation}
\label{eqn:capabilityEffect}
	\bigwedge_{c \in C} c_t \implies \mathit{eff}_c(Q_{t,1})
\end{equation}
The function $\mathit{eff}_c$ returns all output properties and their assured values after the application of capabilities. 

When a capability is applied, changes to directly related properties must be propagated to all synonymous properties. This ensures that all capabilities operate on the same property state. 
\begin{equation}
	\bigwedge_{c \in C} c_t \implies \bigwedge_{q \in \mathit{eff}_c} \bigwedge_{q^{syn} \in Q^{syn}_q} (q_{t,1} =  q^{syn}_{t,1})
\end{equation}
Together, these constraints ensure that if a capability is applied, and thus set to true, all of its preconditions must be met according to its inputs, and all of its effects must be met according to its outputs. 
Otherwise, a capability cannot be applied in that happening. 

\subsubsection{Capability Constraints}
In contrast to the contribution in \cite{CMZ_PlanningforHybridSystems_2020}, expressions between elements of capabilities can be expressed in the form of so-called capability constraints using \emph{OpenMath}. 
These constraints form a mathematical relation between different properties, e.g. product properties, which are used as an input or output of a capability.
For the SMT encoding, these constraints are parsed and corresponding constraints are generated in SMT. 
If a capability constraint refers only to inputs, the constraint is transferred as a precondition for variables of layer $0$ comparable to Equation~\ref{eqn:capabilityPrecondition}. 
If a constraint refers to an output, it is defined as an effect for variables of layer $1$ comparable to Equation~\ref{eqn:capabilityEffect}. 
All other capability constraints are added as an additional SMT constraint.

\subsubsection{Capability Mutexes}
In one happening, multiple capabilities can be applied in parallel. However, if capabilities are mutually exclusive, the encoding needs to ensure that such mutex capabilities cannot be applied in one happening.
We define two capabilities $c_a$ and $c_b$ as mutex if their sets of input or output properties intersect. This is in line with the definition of \cite{CMZ_PlanningforHybridSystems_2020}. 
For two mutex capabilities marked with $\nparallel$, at least one must be false for a given happening. 
\begin{equation}
	\bigwedge_{c \in C} \bigwedge_{\check{c} \in C | c \nparallel \check{c}} (\neg c_t \vee \neg \check{c}_t)
\end{equation}

\subsubsection{Initial and Goal State Description}
So far, equations describing provided capabilities as well as their preconditions and effects on properties have been formulated. 
For a complete encoding of a planning problem, the specific description of the initial state and the desired goal state is also needed. 

In this approach, the initial state is made up of two parts: 
The first part results from the values of properties with an expression goal of \emph{actual value} and associated with provided capabilities or resources. These properties define the current state of the system, for which planning is carried out (e.g., currently equipped tools, part quantities in stock, positions of autonomous robots etc.) before planning.
The second part is formed by the required capability that the current planning task is carried out for. 
All input properties of the required capability with expression goal \emph{requirement} define initial conditions for the planning problem.
If these two parts contradict, i.e., if a required capability requires a certain initial situation that does not correspond with the actual situation, there can be no valid plan for any number of happenings $n$.
Initial values are assigned to the properties in the first happening and in layer $0$, i.e. before the first capability application. 
\begin{equation}
	I(Q_{0,0})
\end{equation}
Finally, the goal state is taken from the required capability and its output properties. These output properties express requirements against desired properties of, for example, a good to be transported, and their values are assigned to the properties in the last happening $n$ and layer $1$, i.e. after the last capability application.  
\begin{equation}
	G(Q_{n,1}) 
\end{equation}

In this way, the properties of required capabilities are encoded.
However, the properties of the required capability must be aligned with their synonymous counterparts of provided capabilities. 
Therefore, the initial and goal states must be set for the synonymous properties, too. 
For this purpose, the sets $pre_{req C}$ and $\mathit{eff}_{req C}$ are used for preconditions and effects of the required capability.  
\begin{equation}
	\bigwedge_{q \in pre_{req C}} \bigwedge_{q^{syn} \in Q^{syn}_q} (q_{0,0} = q^{syn}_{0,0})  
\end{equation}

\begin{equation}
	\bigwedge_{q \in \mathit{eff}_{req C}} \bigwedge_{q^{syn} \in Q^{syn}_q} (q_{n,1} = q^{syn}_{n,1})  
\end{equation}

\subsubsection{Proposition Continuation across Happenings}
Just like in \cite{CMZ_PlanningforHybridSystems_2020}, the following constraints are needed to ensure that boolean variables do not change between happenings.
A proposition in layer $0$ of a happening $t$ is true if the same proposition was already true in layer $1$ of the previous happening $t-1$. 
\begin{equation}
	\bigwedge_{t=1}^n \bigwedge_{p \in P} p_{t,0} \implies p_{t-1,1}
\end{equation}
Likewise, a proposition must be false if it was false in the previous happening. 
\begin{equation}
	\bigwedge_{t=1}^n \bigwedge_{p \in P} \neg p_{t,0} \implies \neg p_{t-1,1}
\end{equation}

\subsubsection{Real Variable Continuation across Happenings}
Unlike in \cite{CMZ_PlanningforHybridSystems_2020}, processes and durative actions are not considered in this paper. Hence, there is no change to the values of real variables between happenings. 
Continuation of real variables thus consists of ensuring that variable values remain the same across happenings. 
\begin{equation}
	\bigwedge_{t=1}^n \bigwedge_{r \in R} r_{t,0} = r_{t-1,1}
\end{equation}
In its current form, the continuation of propositions and real variables across happenings could be treated in the same way.
However, we are currently working on the integration of events and durative actions. With these elements, the continuation of propositions and real variables must be handled separately. 
For this reason, these two types of continuations are listed separately here.

\subsection{Implementation}
\label{subsec:implementation}
We implemented our capability planning approach as a Python application called \emph{CaSkade Planner}\footnote{https://github.com/caskade-automation/caskade-planner} using the Python libraries \emph{rdflib} and \emph{z3-solver} in order to interact with RDF data and create SMT encodings, respectively.
The main algorithm in a high-level schematic is shown in Algorithm~\ref{alg:capsmith}.
In order for any planning to take place at all, there must be an instance of the CaSk ontology $O$ describing both the domain with the available resources and capabilities, as well as the actual problem to be solved with the required capabilities. 
In addition, the upper bound of happenings to consider $n$ is defined. 
The algorithm starts by querying all information about resources, capabilities and properties from the ontology using SPARQL so that all necessary information is available for encoding.
\begin{algorithm}
  \caption{CaSkade Planner}
  \begin{algorithmic}[1]
  \label{alg:capsmith}
    \REQUIRE Capability Ontology $O$, happening bound $n$
    \STATE $input \leftarrow sparqlQueries(O)$; 
    \FOR{$t=0$ to $n$} 
        \STATE $\Pi_t \leftarrow encode(input, t)$;  
        \STATE $solution \leftarrow solve(\Pi_t)$;
        \IF{$solution \models \Pi_t$}
        \STATE $\pi \leftarrow extractPlan(solution)$; 
        \RETURN $\pi$
        \ENDIF
    \ENDFOR
    \RETURN No plan found
  \end{algorithmic}
\end{algorithm}
The queried information is used to generate an encoding as described in Section~\ref{subsec:encoding}.
A loop is used to iterate over the happenings starting at $t=0$ up to the upper bound $n$. 
For every iteration, the encoding is solved using the Z3 solver.
If no solution is found for a happening $t$, the queried information is reused and the encoding is extended by the additional happening.

If the problem is satisfiable, the plan is extracted from the solution. 
For this purpose, the ontology IRIs of all SMT variables are retrieved and the flat SMT variable list is converted into an object structure, in which properties are assigned to their capabilities.
As soon as a solution is found, planning is stopped. 
This means that a found solution always corresponds to the smallest possible number of happenings. 

\section{Conclusion \& Outlook}    
\label{sec:conclusion}
In this contribution, we present an approach that integrates the two aspects of capability modeling and automated planning using SMT. 
The approach eliminates manual efforts for setting up models for automated planning and instead uses machine-interpretable information available in the form of a capability ontology.
In addition, it offers a formal solution for capability planning / capability matching, for which previously there were mostly approaches specialized for individual cases. 
While most of the previous approaches had to be maintained for individual capabilities, our approach is independent of the number of provided capabilities, as it does not define any fixed combination rules between capabilities. 
Instead, capabilities can be automatically combined to form a sequence if the restrictions of their respective inputs and outputs allow it.

In Section \ref{sec:capabilityModeling}, we presented the main elements to model capabilities with the \emph{CaSk} ontology. While the capability aspect is the most important one for the actual planning algorithm, the underlying ontology contains additional information that can be used to invoke executable machine functions using so-called \emph{skills}. 
For the skill aspect, we presented an orchestration mechanism based on \emph{Business Process Model and Notation} (BPMN) in \cite{KVF_ModelingandExecutingProduction_2022}. 
Currently, skill processes must be manually modeled in BPMN. As a future extension of this planning approach, we plan to automatically transform plans into BPMN processes so that they can be executed.

Furthermore, we investigate different ways of providing explanations to human users who have no experience in SMT. When a plan is found, additional information about this plan, e.g., about the capabilities that and the property values it contains, can be retrieved from the ontology. With this information, a planning result can be made more comprehensible to human operators. A similar approach is presented in \cite{GGR_SupportingSkillbasedFlexibleManufacturing_2020}.

In cases where no valid plan can be found, human operators may want understand the reasoning that lead to this decision. In such cases, we plan to use minimal unsatisfiable cores, i.e., the smallest unsatisfiable subset of the original planning clauses, to extract the root cause for unsatisfiability. Enriching elements of an unsatisfiable core with the corresponding information from the ontology seems promising to foster understandability of planning decisions.

\section{Acknowledgments}
This research contribution in the RIVA project is funded by dtec.bw – Digitalization and Technology Research Center of the Bundeswehr. dtec.bw is funded by the European Union – NextGenerationEU.

\bibliography{references}

\end{document}